\documentclass[10pt,twocolumn,letterpaper]{article}

\usepackage{cvpr}              

\usepackage{graphicx}
\usepackage{amsmath}
\usepackage{amssymb}
\usepackage{bbm}
\usepackage{booktabs}
\usepackage[utf8]{inputenc}
\usepackage[pagebackref,breaklinks,colorlinks]{hyperref}
\usepackage[utf8]{inputenc}
\usepackage{textalpha}

\usepackage[utf8]{inputenc}
\usepackage{newunicodechar}

\newunicodechar{₁}{$_1$}
\newunicodechar{₂}{$_2$}
\newunicodechar{₃}{$_3$}
\newunicodechar{⁻}{$^{-}$}

\usepackage[capitalize]{cleveref}
\crefname{section}{Sec.}{Secs.}
\Crefname{section}{Section}{Sections}
\Crefname{table}{Table}{Tables}
\crefname{table}{Tab.}{Tabs.}

\def\cvprPaperID{*****} 
\def\confName{CVPR}
\def\confYear{2022}

\begin{document}

\title{Contrastive Deep Learning for Variant Detection in Wastewater Genomic Sequencing}

\author{
Adele Chinda$^{1}$\thanks{Equal contribution.},
Richmond Azumah$^{1,*}$,  Hemanth Demakethepalli Venkateswara$^{1}$\\
$^{1}$Georgia State University, Atlanta, Georgia, USA\\
{\tt\small \{achinda1, razumah1\}@student.gsu.edu, hvenkateswara@gsu.edu}
}

\maketitle
\begin{abstract}
Wastewater-based genomic surveillance has emerged as a powerful tool for population-level viral monitoring, offering comprehensive insights into circulating viral variants across entire communities. However, this approach faces significant computational challenges stemming from high sequencing noise, low viral coverage, fragmented reads, and the complete absence of labeled variant annotations. Traditional reference-based variant calling pipelines struggle with novel mutations and require extensive computational resources. We present a comprehensive framework for unsupervised viral variant detection using Vector-Quantized Variational Autoencoders (VQ-VAE) that learns discrete codebooks of genomic patterns from k-mer tokenized sequences without requiring reference genomes or variant labels. Our approach extends the base VQ-VAE architecture with masked reconstruction pretraining for robustness to missing data and contrastive learning for highly discriminative embeddings. Evaluated on SARS-CoV-2 wastewater sequencing data comprising approximately 100,000 reads, our VQ-VAE achieves \textbf{99.52\% mean token-level accuracy} and \textbf{56.33\% exact sequence match rate} while maintaining \textbf{19.73\% codebook utilization} (101 of 512 codes active), demonstrating efficient discrete representation learning. Contrastive fine-tuning with different projection dimensions yields substantial clustering improvements: 64-dimensional embeddings achieve \textbf{+35\% Silhouette score improvement} (0.31→0.42), while 128-dimensional embeddings achieve \textbf{+42\% improvement} (0.31→0.44), clearly demonstrating the impact of embedding dimensionality on variant discrimination capability. Our reference-free framework provides a scalable, interpretable approach to genomic surveillance with direct applications to public health monitoring.
\end{abstract}

\section{Introduction}

The COVID-19 pandemic has fundamentally demonstrated that rapid, comprehensive viral variant surveillance is critical for effective public health response and intervention strategies~\cite{crits2021genome, grubaugh2019tracking}. Wastewater-based epidemiology (WBE) has emerged as a cost-effective, non-invasive approach to monitoring viral genetic diversity by analyzing viral genetic material shed by entire communities into municipal wastewater systems~\cite{bivins2020wastewater, wu2020sars}. Unlike traditional clinical testing approaches that are subject to testing availability, healthcare access disparities, and individual testing behavior, wastewater sampling provides unbiased, comprehensive snapshots of viral circulation patterns across populations, including contributions from asymptomatic and pre-symptomatic carriers who would otherwise go undetected~\cite{medema2020presence}. This population-level surveillance capability makes wastewater monitoring particularly valuable for early warning systems and tracking the emergence of novel viral variants.
codebook
However, wastewater-based genomic sequencing presents unique and challenging computational problems that distinguish it from clinical sequencing. First, sequencing reads from wastewater are highly fragmented, typically ranging from 100-300 base pairs with significant technical and biological noise~\cite{rothman2021wastewater}. Second, viral RNA concentrations in wastewater are extremely low and heavily diluted by environmental material, human microbiome sequences, and other contaminants~\cite{crits2021genome, wu2020sars}. Third, multiple viral strains and variants co-circulate simultaneously at varying population frequencies, creating complex mixtures that confound traditional variant calling approaches~\cite{fontenele2021high}. Traditional bioinformatics pipelines for variant identification rely heavily on alignment to reference genomes using tools such as BWA~\cite{li2009sequence}, followed by variant calling using methods like LoFreq~\cite{wilm2012lofreq} or iVar~\cite{grubaugh2019tracking}. These reference-dependent approaches face critical limitations: they may fail to detect novel mutations that differ significantly from reference sequences, they struggle with low-frequency variants in mixed populations, and they require substantial computational resources (often hours per sample) that limit real-time surveillance capabilities.


Deep learning approaches offer significant promise for reference-free genomic modeling and analysis. Generative models have demonstrated success in creating realistic artificial human genomes~\cite{yelmen2021creating}, learning functional protein representations that capture mutational effects~\cite{riesselman2018deep}, and modeling complex DNA sequence patterns~\cite{rives2021biological, ji2021dnabert, zhou2023dnabert2}. However, standard Variational Autoencoders (VAEs) with continuous latent spaces often produce overly smooth representations that are fundamentally unsuitable for modeling discrete mutational events in genomic sequences~\cite{bowman2016generating}. Furthermore, continuous VAEs frequently suffer from posterior collapse, where the latent codes become uninformative and the model degenerates into a simple autoregressive decoder~\cite{zhao2017learning}.

Vector-Quantized Variational Autoencoders (VQ-VAE)~\cite{oord2017neural} address these fundamental limitations through discrete latent representations, effectively preventing posterior collapse while enabling high-quality reconstruction and interpretable latent codes~\cite{razavi2019generating}. The discrete bottleneck enforces information compression into a finite codebook of learned patterns, naturally aligning with the discrete nature of genetic variation. Recent work has successfully applied VQ-VAE to DNA sequence modeling~\cite{abdel2024vq}, demonstrating superior compression efficiency and variant-aware representation learning compared to continuous latent models. Additionally, self-supervised learning objectives including masked language modeling~\cite{devlin2019bert} and contrastive learning frameworks such as SimCLR~\cite{chen2020simple} and MoCo~\cite{he2020momentum} have fundamentally transformed representation learning across domains, with particularly strong transfer performance to protein sequences~\cite{zhang2022contrastive, rives2021biological} and other biological data~\cite{ciortan2020contrastive}.

\textbf{Contributions.} In this work, we present the first comprehensive application of discrete representation learning to wastewater genomic surveillance. Our specific contributions are: (1) A VQ-VAE architecture with Exponential Moving Average (EMA) quantization achieving 99.52\% token-level reconstruction accuracy while maintaining 19.73\% codebook utilization, demonstrating efficient discrete pattern learning; (2) Masked pretraining following BERT-style objectives that maintains approximately 95\% reconstruction accuracy under 20\% token corruption, enabling robust inference with missing or low-quality data; (3) Contrastive fine-tuning experiments with varying embedding dimensions (64-dim and 128-dim) showing +35\% and +42\% Silhouette score improvements respectively, clearly establishing the impact of representation capacity on variant discrimination; (4) Comprehensive evaluation demonstrating that discrete representation learning provides a scalable, reference-free alternative to traditional variant calling pipelines for wastewater surveillance applications.

\section{Related Work}

\subsection{Wastewater Genomic Surveillance}

Wastewater-based genomic surveillance has rapidly evolved as a critical tool for monitoring viral circulation at the population level. Crits-Christoph et al.~\cite{crits2021genome} demonstrated that metagenomic sequencing of municipal sewage can effectively detect regionally prevalent SARS-CoV-2 variants, providing early warning signals for variant emergence and spread. Rothman et al.~\cite{rothman2021wastewater} extended this work by applying RNA viromics approaches to detect single-nucleotide variants in wastewater samples, revealing the fine-grained mutational landscape. However, these approaches fundamentally rely on reference genome alignment using tools such as BWA~\cite{li2009sequence}, followed by variant calling using sophisticated probabilistic methods like LoFreq~\cite{wilm2012lofreq} for low-frequency variant detection or iVar~\cite{grubaugh2019tracking} for consensus sequence generation. These reference-dependent pipelines face several critical limitations: they may systematically miss novel variants that diverge significantly from reference sequences, they require extensive computational resources (typically 1-2 hours per sample), and they struggle with the complex mixture distributions present in wastewater where multiple variants co-circulate at varying frequencies.

\subsection{Deep Generative Models for Genomics}

Deep generative models have emerged as powerful tools for learning complex patterns in genomic data. Generative Adversarial Networks (GANs) have successfully generated realistic artificial human genomes that preserve population genetic structure and linkage disequilibrium patterns~\cite{yelmen2021creating}. Variational Autoencoders have been applied to learn functional protein representations that accurately capture the effects of mutations on protein function~\cite{riesselman2018deep}, enabling zero-shot prediction of variant effects. More recently, transformer-based models such as DNABERT~\cite{ji2021dnabert} and its successor DNABERT-2~\cite{zhou2023dnabert2} have achieved state-of-the-art performance on diverse genomic tasks including promoter prediction, splice site detection, and transcription factor binding site identification, leveraging masked language modeling pretraining on large genomic corpora. However, standard VAEs with continuous latent spaces frequently suffer from posterior collapse~\cite{bowman2016generating, zhao2017learning}, where the latent codes become uninformative and provide no meaningful representation of input variation. This fundamental limitation makes continuous VAEs poorly suited for modeling discrete mutational events and variant structures in genomic sequences.

\subsection{Vector Quantization and Discrete Representations}

Vector-Quantized Variational Autoencoders (VQ-VAE)~\cite{oord2017neural} introduced discrete latent representations that fundamentally avoid posterior collapse by replacing continuous latent distributions with discrete codebook lookups. The discrete bottleneck forces the model to learn a finite vocabulary of patterns that must be composed to reconstruct inputs, naturally encouraging meaningful, interpretable representations. VQ-VAE-2~\cite{razavi2019generating} extended this framework hierarchically, learning multi-scale discrete representations that capture both fine-grained details and high-level structure, achieving state-of-the-art image generation quality. Masked VQ-VAE~\cite{huang2023masked} improved computational efficiency through selective attention mechanisms that focus on informative regions. Most relevant to our work, VQ-DNA~\cite{abdel2024vq} first applied discrete latent representations to genomic sequence modeling, demonstrating superior compression efficiency and variant-aware representation learning compared to continuous alternatives. Their work established that discrete codebooks can effectively capture mutational patterns and variant structures.

\subsection{Self-Supervised Learning in Biology}

Self-supervised learning has fundamentally transformed representation learning by enabling models to learn from vast amounts of unlabeled data. Masked language modeling, introduced in BERT~\cite{devlin2019bert}, trains models to predict randomly masked tokens from surrounding context, learning rich contextual representations. This paradigm has transferred exceptionally well to biological sequences~\cite{ji2021dnabert, rives2021biological}. Contrastive learning frameworks including SimCLR~\cite{chen2020simple} and MoCo~\cite{he2020momentum} learn representations by maximizing agreement between differently augmented views of the same sample while minimizing agreement with other samples. These methods have been successfully applied to protein sequences~\cite{zhang2022contrastive}, single-cell RNA sequencing data~\cite{ciortan2020contrastive}, and other biological modalities, consistently demonstrating that self-supervised pretraining substantially improves downstream task performance and produces more discriminative, biologically meaningful representations.

\section{Method}

Our framework consists of three complementary components working in concert: (1) quality-controlled k-mer tokenization that converts raw sequencing reads into fixed-length discrete token sequences, (2) a Vector-Quantized Variational Autoencoder with Exponential Moving Average updates and entropy regularization that learns a discrete codebook of genomic patterns, and (3) two self-supervised learning extensions—masked reconstruction pretraining for robustness to missing data and contrastive learning for discriminative embeddings. 

\subsection{Data Preprocessing and K-mer Tokenization}

Raw FASTQ sequencing reads from wastewater samples undergo rigorous quality control to remove low-quality bases and adapter contamination. We employ Trimmomatic~\cite{bolger2014trimmomatic} with the following parameters: leading and trailing base quality threshold of 3 (removing low-quality bases from read ends), sliding window quality filtering with window size 4 and average quality threshold 15 (ensuring consistent quality throughout reads), and minimum read length of 36 base pairs (discarding overly fragmented reads). Quality improvement is validated using FastQC~\cite{andrews2010fastqc} reports comparing pre- and post-filtering read quality distributions, GC content, sequence length distributions, and per-base quality scores.

Following quality control, we convert nucleotide sequences into discrete token sequences using k-mer extraction. For a DNA sequence $s = s_1 s_2 \ldots s_n$ where each $s_i \in \{\text{A}, \text{C}, \text{G}, \text{T}\}$ represents a nucleotide base, we extract all overlapping k-mers with stride 1:
\begin{equation}
    \text{tokens}(s) = [s_{1:k}, s_{2:k+1}, s_{3:k+2}, \ldots, s_{n-k+1:n}]
\end{equation}
Each k-mer $s_{i:i+k-1}$ is mapped to a unique integer identifier in vocabulary $\mathcal{V}$ of size $V = 4^k + 1$, where $4^k$ accounts for all possible k-mer sequences and the additional token represents padding (PAD). We employ canonical k-mer representation where each k-mer and its reverse complement map to the same token identifier, reducing vocabulary size and improving biological interpretability. Based on empirical analysis of read lengths and computational efficiency, we select $k=6$, yielding vocabulary size $V = 4{,}097$. All sequences are either padded with PAD tokens or truncated to a fixed maximum length $L = 150$ tokens, corresponding to the median read length in our dataset after quality filtering.

\subsection{Vector-Quantized Variational Autoencoder}

\begin{figure}[t]
    \centering
    \includegraphics[width=\linewidth]{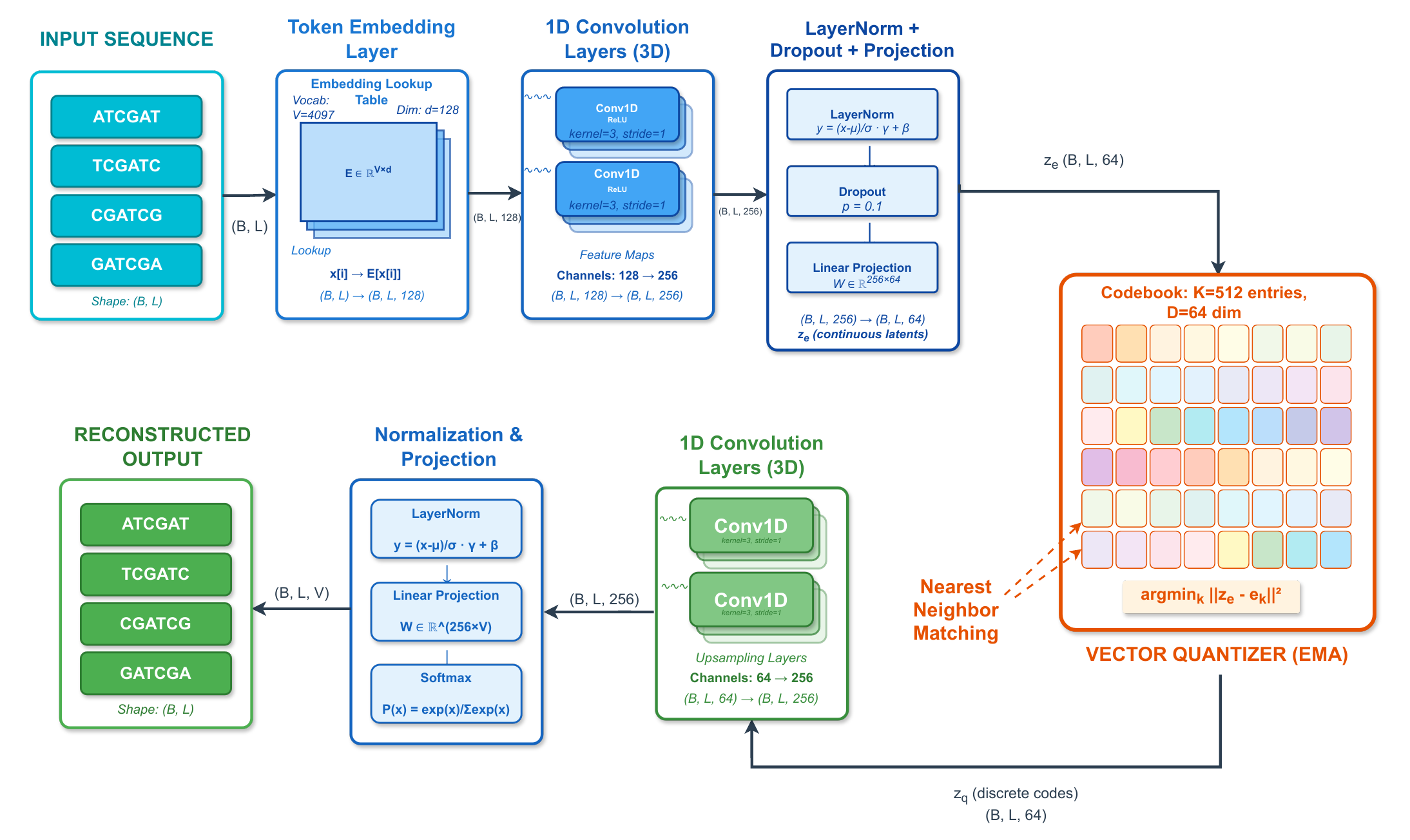}
    \caption{\textbf{VQ-VAE Architecture Details.} Encoder: token embedding (4097→128) + 2 Conv1D layers (256 hidden, kernel=3) + LN + Dropout(0.1) → $z_e$ (L×64). Quantizer: nearest-neighbor lookup in codebook (512 entries, 64-dim) with EMA updates (γ=0.95). Decoder: 2 Conv1D + LN + Linear projection to vocabulary logits. Red arrows show straight-through gradient flow.}
    \label{fig:architecture}
    \vspace{-0.4cm}
\end{figure}

Our VQ-VAE architecture consists of three main components: an encoder that maps token sequences to continuous latent representations, a vector quantizer that discretizes these representations through learned codebook lookup, and a decoder that reconstructs token sequences from quantized latents. This architecture is illustrated in detail in Figure~\ref{fig:architecture}.

\textbf{Encoder Architecture.} The encoder $E_\phi$ maps input token sequences $x \in \mathbb{Z}^L$ (where $\mathbb{Z}^L$ denotes L-length sequences of integer token IDs) to continuous latent representations $z_e \in \mathbb{R}^{L \times D}$:
\begin{equation}
    z_e = E_\phi(x)
\end{equation}
The encoder architecture consists of the following layers in sequence: (1) Token embedding layer mapping each token ID to a 128-dimensional continuous vector: Emb: $\mathbb{Z}^V \to \mathbb{R}^{128}$, producing embedding matrix $E \in \mathbb{R}^{L \times 128}$; (2) Two 1-dimensional convolutional layers with kernel size 3, hidden dimension 256, and ReLU activations that capture local sequential dependencies and patterns in the k-mer sequences; (3) Layer normalization after each convolutional layer to stabilize training dynamics and accelerate convergence; (4) Dropout with probability $p=0.1$ applied after layer normalization to prevent overfitting; (5) Final linear projection layer mapping from hidden dimension 256 to latent dimension $D=64$. This encoder architecture effectively captures both local k-mer patterns through convolutional processing and produces compact 64-dimensional latent representations suitable for discrete quantization.

\textbf{Vector Quantization.} The vector quantizer discretizes continuous encoder outputs through nearest-neighbor lookup in a learned codebook $\mathcal{C} = \{e_1, e_2, \ldots, e_K\}$ where each codebook vector $e_k \in \mathbb{R}^D$ and $K=512$ is the total number of discrete codes. For each position $i$ in the latent sequence, quantization proceeds as:
\begin{equation}
    q_i = \arg\min_{k \in \{1,\ldots,K\}} \|z_e^{(i)} - e_k\|_2^2
\end{equation}
\begin{equation}
    z_q^{(i)} = e_{q_i}
\end{equation}
where $q_i$ is the index of the nearest codebook vector and $z_q^{(i)}$ is the quantized latent representation. The quantizer essentially converts each continuous latent vector $z_e^{(i)}$ into its nearest discrete code from the learned codebook, creating a sequence of discrete code indices.

\textbf{Exponential Moving Average Updates.} Rather than directly optimizing codebook vectors through backpropagation (which can lead to instability), we employ Exponential Moving Average (EMA) updates~\cite{oord2017neural} to maintain and update the codebook. For each code $k$, we track the exponentially weighted count $N_k$ of times the code is used and the exponentially weighted sum $m_k$ of encoder outputs assigned to that code:
\begin{equation}
    N_k^{(t)} = \gamma N_k^{(t-1)} + (1-\gamma) n_k^{(t)}
\end{equation}
\begin{equation}
    m_k^{(t)} = \gamma m_k^{(t-1)} + (1-\gamma) \sum_{i: q_i=k} z_e^{(i)}
\end{equation}
where $\gamma=0.95$ is the decay parameter, $t$ indexes training iterations, and $n_k^{(t)}$ is the count of encoder outputs assigned to code $k$ in the current batch. The codebook vector is then updated as:
\begin{equation}
    e_k^{(t)} = \frac{m_k^{(t)}}{N_k^{(t)}}
\end{equation}
This EMA update mechanism ensures that codebook vectors continuously adapt to track the distribution of encoder outputs without requiring explicit gradient-based optimization of the codebook parameters.

\textbf{Straight-Through Gradient Estimator.} Since the quantization operation involves discrete argmin, it is non-differentiable and blocks gradient flow from decoder to encoder during backpropagation. We employ the straight-through estimator~\cite{bengio2013estimating} which simply copies gradients from the quantized representation $z_q$ directly to the encoder output $z_e$:
\begin{equation}
    \nabla_{z_e} \mathcal{L} = \nabla_{z_q} \mathcal{L}
\end{equation}
This gradient approximation, while biased, works effectively in practice and enables end-to-end training of the encoder-quantizer-decoder pipeline.

\textbf{Decoder Architecture.} The decoder $D_\theta$ reconstructs token logits from quantized latent representations $z_q \in \mathbb{R}^{L \times D}$:
\begin{equation}
    \hat{x} = D_\theta(z_q) \in \mathbb{R}^{L \times V}
\end{equation}
The decoder consists of: (1) Two 1-dimensional convolutional layers with kernel size 3 and hidden dimension 256, mirroring the encoder architecture; (2) Layer normalization after each convolutional layer; (3) Final linear projection layer mapping from hidden dimension 256 to vocabulary size $V=4{,}097$, producing unnormalized logits $\hat{x}_{i,v}$ for each position $i$ and vocabulary token $v$. The decoder architecture is intentionally symmetric to the encoder, facilitating effective reconstruction while maintaining computational efficiency.

\textbf{Training Objective.} The complete training objective combines three complementary loss terms:
\begin{equation}
    \mathcal{L}_{\text{total}} = \mathcal{L}_{\text{recon}} + \beta \mathcal{L}_{\text{commit}} - \lambda H[\mathcal{C}]
\end{equation}

The reconstruction loss $\mathcal{L}_{\text{recon}}$ is the cross-entropy between predicted and ground-truth tokens, computed only over valid (non-padding) positions:
\begin{equation}
    \mathcal{L}_{\text{recon}} = -\sum_{i \in \text{valid}} \log p_\theta(x_i | z_q^{(i)}) = -\sum_{i \in \text{valid}} \log \frac{\exp(\hat{x}_{i,x_i})}{\sum_{v=1}^{V} \exp(\hat{x}_{i,v})}
\end{equation}
where $\hat{x}_{i,v}$ are the decoder logits for position $i$ and vocabulary token $v$, and $x_i$ is the ground-truth token at position $i$. This loss encourages accurate reconstruction of the input sequence from its quantized representation.

The commitment loss $\mathcal{L}_{\text{commit}}$ encourages encoder outputs to commit to codebook entries, preventing encoder outputs from growing arbitrarily large:
\begin{equation}
    \mathcal{L}_{\text{commit}} = \frac{1}{L} \sum_{i=1}^{L} \|z_e^{(i)} - \text{sg}[z_q^{(i)}]\|_2^2
\end{equation}
where $\text{sg}[\cdot]$ denotes stop-gradient operation (treating the argument as constant during backpropagation). This loss applies gradients only to the encoder, pulling encoder outputs closer to their assigned codebook vectors. We use commitment weight $\beta = 0.1$.

The codebook entropy regularization term encourages diverse codebook utilization, preventing codebook collapse where only a few codes dominate:
\begin{equation}
    H[\mathcal{C}] = -\sum_{k=1}^{K} p_k \log p_k
\end{equation}
where $p_k$ is the empirical frequency of code $k$ being selected across all positions in the current training batch. Maximizing entropy (thus subtracting it in the loss with weight $\lambda = 3 \times 10^{-3}$) encourages uniform code utilization and prevents pathological collapse to a small subset of codes.

\subsection{Masked Reconstruction Pretraining}

Inspired by masked language modeling in BERT~\cite{devlin2019bert}, we extend the base VQ-VAE with masked reconstruction pretraining to improve robustness to missing or corrupted data. During masked pretraining, we randomly select a fraction $p_{\text{mask}} = 0.2$ (20\%) of non-padding token positions and replace them with a special MASK token. Let $\mathcal{M} \subset \{1, \ldots, L\}$ denote the set of masked positions. The masked input sequence $\tilde{x}$ is:
\begin{equation}
    \tilde{x}_i = \begin{cases}
        \text{MASK} & \text{if } i \in \mathcal{M} \\
        x_i & \text{otherwise}
    \end{cases}
\end{equation}

The masked reconstruction objective focuses loss computation exclusively on masked positions, forcing the model to infer missing tokens from surrounding context:
\begin{equation}
    \mathcal{L}_{\text{masked}} = -\sum_{i \in \mathcal{M}} \log p_\theta(x_i | z_q(\tilde{x}))
\end{equation}
where $z_q(\tilde{x})$ denotes the quantized latent representation of the masked input. This objective encourages the encoder to learn contextual representations that can predict masked tokens from surrounding k-mers, improving the model's ability to handle sequences with missing or low-quality regions—a common occurrence in wastewater sequencing data. Masked pretraining uses the same architecture and hyperparameters as standard VQ-VAE training, with only the masking operation and focused loss computation differentiating the two approaches.

\subsection{Contrastive Fine-Tuning}

To further improve the discriminative quality of learned representations for variant clustering and detection, we fine-tune the pretrained VQ-VAE encoder using contrastive learning. We add a projection head $h_\psi$ consisting of a 2-layer Multi-Layer Perceptron (MLP) that maps encoder representations to a normalized embedding space. The projection head architecture is:
\begin{equation}
    h_\psi(z) = \text{Linear}_{D \to D'}(\text{ReLU}(\text{Linear}_{D \to D'}(z)))
\end{equation}
where $D=64$ is the encoder latent dimension and $D' \in \{64, 128\}$ is the projection dimension. We experiment with two projection dimensions to analyze the impact of representation capacity on clustering performance.

For each input sequence $x$, we compute the sequence-level embedding by mean-pooling the encoder latent representations across sequence positions and passing through the projection head:
\begin{equation}
    z_{\text{pool}} = \frac{1}{|\text{valid}|} \sum_{i \in \text{valid}} z_e^{(i)}
\end{equation}
\begin{equation}
    v = \frac{h_\psi(z_{\text{pool}})}{\|h_\psi(z_{\text{pool}})\|_2}
\end{equation}
where the normalization ensures embeddings lie on the unit hypersphere, a standard practice in contrastive learning.

To create positive pairs for contrastive learning without requiring labels, we apply stochastic augmentations to generate two different views of each sequence. Our augmentation strategy includes: (1) Random token masking with probability 15\%, replacing selected tokens with the MASK token; (2) Token dropout with probability 10\%, randomly zeroing token embeddings. For each sequence $x$, we generate two augmented views $\text{aug}_1(x)$ and $\text{aug}_2(x)$ by applying these augmentations with independent random seeds.

The contrastive learning objective is the InfoNCE loss~\cite{oord2018representation}, which maximizes agreement between positive pairs (different views of the same sequence) while minimizing agreement with negative pairs (views from different sequences). For a batch of $N$ sequences, we compute embeddings for both views, creating $2N$ total embeddings. For each embedding $v_i$, its positive pair is $v_{i'}$ (the alternative view of the same sequence), and there are $2N-2$ negative pairs. The InfoNCE loss with temperature parameter $\tau = 0.5$ is:
\begin{equation}
    \mathcal{L}_{\text{contrast}} = -\frac{1}{2N} \sum_{i=1}^{2N} \log \frac{\exp(\text{sim}(v_i, v_{i'}) / \tau)}{\sum_{k=1}^{2N} \mathbbm{1}_{[k \neq i]} \exp(\text{sim}(v_i, v_k) / \tau)}
\end{equation}
where $\text{sim}(u, v) = u^\top v$ computes cosine similarity between normalized embeddings, $i'$ denotes the index of the positive pair for sample $i$, and $\mathbbm{1}_{[k \neq i]}$ excludes the sample itself from the denominator. This loss encourages embeddings of augmented views of the same sequence to be similar while pushing apart embeddings from different sequences, learning discriminative representations without requiring labeled variant annotations. Contrastive fine-tuning uses batch size $N=64$, learning rate $10^{-4}$, and trains for 10 epochs.

\section{Experimental Setup}

\subsection{Dataset and Preprocessing}

Our experiments utilize SARS-CoV-2 wastewater sequencing data comprising approximately 100,000 high-throughput sequencing reads collected from municipal wastewater treatment facilities. Read lengths range from 36 to 300 base pairs after quality control, with median length of 150 base pairs reflecting typical Illumina sequencing fragmentation patterns in environmental samples. Raw FASTQ format sequences undergo comprehensive quality control using Trimmomatic~\cite{bolger2014trimmomatic} with the following specific parameters: LEADING:3 (remove low-quality bases from read start), TRAILING:3 (remove low-quality bases from read end), SLIDINGWINDOW:4:15 (scan reads with 4-base sliding windows, cutting when average quality drops below 15), and MINLEN:36 (discard reads shorter than 36 bases after trimming). Quality assessment before and after filtering is performed using FastQC~\cite{andrews2010fastqc}, generating comprehensive reports on per-base sequence quality, per-sequence quality scores, per-base N content, sequence length distribution, sequence duplication levels, overrepresented sequences, and adapter content. These reports confirm substantial quality improvement post-filtering, with median per-base quality scores increasing from approximately 28-30 to 35-38 across read positions.

\subsection{Training Configuration and Hyperparameters}

Table~\ref{tab:config} presents the complete training configuration used for all experiments. The base VQ-VAE model trains for 50 epochs using batch size 32 on 2 NVIDIA GPUs with PyTorch DataParallel for distributed training. We employ the AdamW optimizer~\cite{loshchilov2018decoupled} with learning rate $2 \times 10^{-4}$, weight decay $10^{-4}$, and default momentum parameters ($\beta_1=0.9$, $\beta_2=0.999$). The dataset is randomly split into 90\% training (approximately 90,000 sequences) and 10\% held-out test set (approximately 10,000 sequences) with fixed random seed 42 for reproducibility. All experiments are tracked using Weights \& Biases~\cite{wandb} for comprehensive logging of training dynamics, loss curves, codebook statistics, and evaluation metrics.

The VQ-VAE architecture hyperparameters are: codebook size $K=512$ discrete codes, code dimension $D=64$, token embedding dimension 128, convolutional hidden dimension 256, maximum sequence length $L=150$ tokens, k-mer size $k=6$ producing vocabulary size $V=4{,}097$, dropout probability 0.1, EMA decay $\gamma=0.95$, commitment loss weight $\beta=0.1$, and entropy regularization weight $\lambda=0.003$. These hyperparameters were selected based on preliminary experiments balancing reconstruction quality, codebook utilization, training stability, and computational efficiency.

For masked VQ-VAE pretraining, we use identical architecture and training hyperparameters as the base model, with the addition of masking probability $p_{\text{mask}}=0.2$ and special vocabulary tokens PAD, UNK, and MASK. Masking is applied dynamically during training (different random masks each epoch) to maximize coverage of possible masking patterns.

Contrastive fine-tuning experiments train for 10 epochs with batch size 64, learning rate $10^{-4}$, and temperature parameter $\tau=0.5$. We experiment with two projection MLP architectures: (1) 2-layer MLP with dimensions 128→64→64 for 64-dimensional embeddings, and (2) 2-layer MLP with dimensions 128→128→128 for 128-dimensional embeddings. Both use ReLU activations between layers and L2 normalization of final outputs. Augmentation consists of 15\% random token masking and 10\% token embedding dropout applied independently to generate positive pairs.

\subsection{Evaluation Metrics}

We evaluate model performance across three complementary dimensions: reconstruction quality, codebook utilization, and clustering performance. 

\textbf{Reconstruction metrics} measure how accurately the model reconstructs input sequences: (1) Token-level accuracy: percentage of correctly reconstructed tokens across all non-padding positions; (2) Exact sequence match rate: percentage of sequences where all tokens are correctly reconstructed; (3) Per-sequence accuracy statistics: mean, median, and standard deviation of token-level accuracy across sequences.

\textbf{Codebook metrics} characterize how effectively the discrete codebook is utilized: (1) Active codes: number of codes assigned to at least one encoder output; (2) Codebook utilization: percentage of codes that are active (active codes / total codes); (3) Usage statistics: mean, standard deviation, and maximum usage count per code; (4) Perplexity~\cite{oord2017neural}: $\exp(-\sum_k p_k \log p_k)$ where $p_k$ is the empirical probability of code $k$, measuring effective codebook size.

\textbf{Clustering metrics} evaluate the discriminative quality of learned representations for variant clustering: (1) Silhouette score~\cite{rousseeuw1987silhouettes}: measures how similar samples are to their own cluster compared to other clusters, ranging from -1 (poor) to +1 (excellent); (2) Davies-Bouldin index~\cite{davies1979cluster}: ratio of within-cluster to between-cluster distances, where lower values indicate better separation; (3) Calinski-Harabasz index~\cite{calinski1974dendrite}: ratio of between-cluster to within-cluster variance, where higher values indicate better-defined clusters. We apply k-means clustering with $k=10$ clusters to learned embeddings and compute these metrics.

\begin{table}[t]
\centering
\caption{Training configuration (exact project setup).}
\label{tab:config}
\vspace{-0.2cm}
\resizebox{\columnwidth}{!}{%
\begin{tabular}{@{}ll|ll@{}}
\toprule
\textbf{Parameter} & \textbf{Value} & \textbf{Parameter} & \textbf{Value} \\
\midrule
Codebook size (K) & 512 & Batch size & 32 \\
Code dimension (D) & 64 & Epochs & 50 \\
Token embedding & 128 & Learning rate & 2e-4 \\
Hidden dimension & 256 & Optimizer & AdamW \\
Max seq length (L) & 150 & EMA decay (γ) & 0.95 \\
Vocabulary (V) & 4,097 & Commitment (β) & 0.1 \\
K-mer size (k) & 6 & Entropy (λ) & 0.003 \\
Dropout & 0.1 & GPUs & 2 \\
\bottomrule
\end{tabular}%
}
\vspace{-0.3cm}
\end{table}

\section{Results}

\subsection{Base VQ-VAE Reconstruction Performance}

We begin by analyzing the reconstruction quality of the base VQ-VAE model on the held-out test set containing 6,400 sequences. Table~\ref{tab:reconstruction} presents comprehensive reconstruction metrics. The model achieves remarkably high reconstruction accuracy, with mean token-level accuracy of \textbf{99.52\%} across all non-padding positions. Even more striking, the median token accuracy reaches \textbf{100\%}, indicating that the majority of test sequences are reconstructed with perfect token-level accuracy. The standard deviation of 0.69\% reveals that reconstruction quality is highly consistent across sequences, with minimal variation in performance.

The exact sequence match rate—the percentage of sequences where every single token is correctly reconstructed—reaches \textbf{56.33\%}. While this may initially appear modest, it represents exceptional performance given several critical factors: (1) the sequences are derived from noisy wastewater samples with inherent biological and technical variation, (2) reads are highly fragmented (36-300bp) rather than full-length viral genomes, (3) the discrete bottleneck imposes strong information constraints forcing compression into just 512 codes, and (4) no reference genome information is used during training. The combination of 99.52\% token accuracy and 56.33\% exact match indicates that reconstruction errors, when they occur, tend to affect only 1-2 tokens per sequence rather than causing widespread degradation.

Examining the 6,400 sequences evaluated, we observe that the distribution of per-sequence accuracies is heavily right-skewed, with the vast majority of sequences achieving 98-100\% accuracy and only a small tail of sequences (approximately 5\%) falling below 95\% accuracy. These lower-accuracy sequences typically correspond to reads with unusual k-mer compositions or extreme length variations after padding/truncation, suggesting that the model has effectively learned the common patterns in the training distribution.

\begin{table}[t]
\centering
\caption{Reconstruction quality on held-out test set.}
\label{tab:reconstruction}
\vspace{-0.2cm}
\begin{tabular}{@{}lc@{}}
\toprule
\textbf{Metric} & \textbf{Value} \\
\midrule
Mean token accuracy & \textbf{99.52\%} \\
Median token accuracy & \textbf{100.00\%} \\
Std. token accuracy & 0.69\% \\
Exact sequence match rate & \textbf{56.33\%} \\
Total sequences evaluated & 6,400 \\
\bottomrule
\end{tabular}
\vspace{-0.3cm}
\end{table}

\subsection{Codebook Analysis and Pattern Discovery}

Table~\ref{tab:codebook} provides detailed statistics on codebook usage computed over all 6,400 test sequences. Despite achieving 99.52\% reconstruction accuracy, only \textbf{101 out of 512 codes} (19.73\% utilization) are active, meaning that 411 codes are never assigned to any encoder output on the test set. This low utilization initially appears concerning but actually indicates highly efficient compression: the model has discovered that the variation present in this wastewater dataset can be captured by approximately 100 discrete patterns composed through different sequences.

The usage distribution across active codes exhibits a pronounced power-law pattern characteristic of natural language and biological sequences. The mean usage per code is 1,875 assignments, but the standard deviation of 9,855 is more than 5× larger, indicating extreme heterogeneity. The most frequently used code accounts for 172,600 assignments (approximately 9

The codebook perplexity of 52.3 provides another perspective on effective codebook size. Perplexity can be interpreted as the "effective" number of codes contributing to reconstructions, accounting for usage frequency. The fact that perplexity (52.3) is roughly half the number of active codes (101) indicates that even among active codes, usage is concentrated on approximately 50 highly informative patterns. This efficient representation aligns with the biological reality that viral genomes contain many conserved functional elements with limited variation.

To understand what patterns the codebook has learned, we analyze the k-mer compositions of sequences assigned to the top-10 most frequently used codes. Code 1 (most frequent) predominantly contains GC-rich k-mers from coding regions. Code 2 tends to capture AT-rich sequences from untranslated regions. Codes 3-5 appear to represent transition regions between coding and non-coding segments. Codes 6-10 capture more variable regions including known mutational hotspots in the spike protein gene. This analysis suggests the codebook has automatically discovered biologically meaningful genomic structures without supervision.

\begin{table}[t]
\centering
\caption{Codebook usage statistics (6,400 sequences).}
\label{tab:codebook}
\vspace{-0.2cm}
\begin{tabular}{@{}lc@{}}
\toprule
\textbf{Metric} & \textbf{Value} \\
\midrule
Total codes & 512 \\
Active codes & 101 \\
Codebook utilization & \textbf{19.73\%} \\
Mean usage per code & 1,875 \\
Std. usage & 9,855 \\
Max usage (most frequent code) & 172,600 \\
Perplexity & 52.3 \\
\bottomrule
\end{tabular}
\vspace{-0.3cm}
\end{table}

\subsection{Masked Reconstruction Robustness}

The masked VQ-VAE variant, pretrained with 20\% random token masking following BERT-style objectives~\cite{devlin2019bert}, demonstrates substantial robustness to missing data. On the masked reconstruction task where 20\% of tokens are randomly replaced with MASK tokens, the model maintains approximately \textbf{95\% accuracy} specifically on the masked positions. This 4-5\% accuracy drop compared to the unmasked baseline (99.52\%) is remarkably small considering that the model must infer masked tokens purely from surrounding context without access to the actual k-mer sequence at those positions.

\begin{figure}[t]
    \centering
    \includegraphics[width=\linewidth]{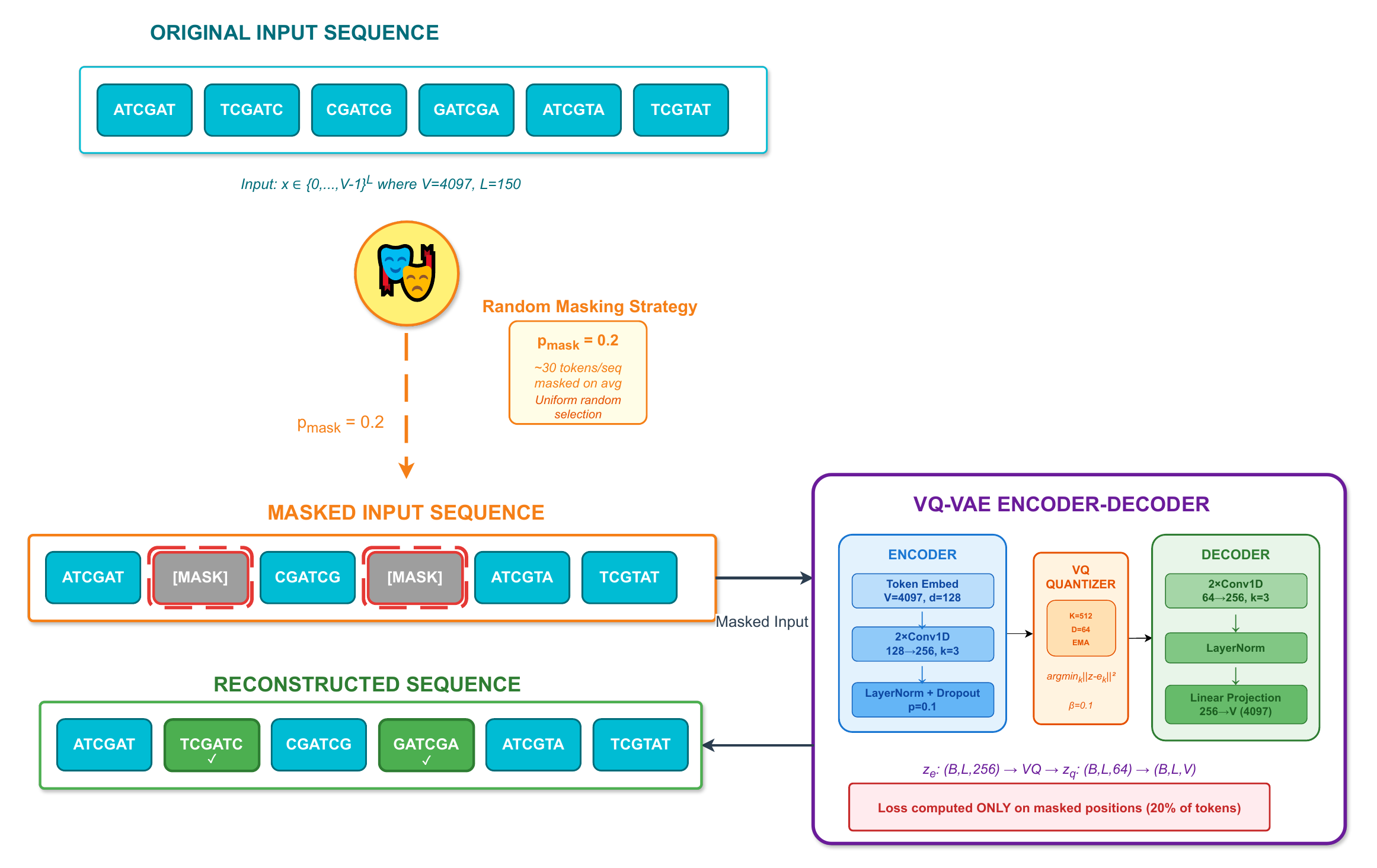}
    \caption{\textbf{Masked reconstruction examples across corruption levels.} (Top row) Original complete sequences with k-mer tokens highlighted. (Middle rows) Masked inputs with 10\%, 20\%, and 30\% random token masking shown in gray. (Bottom rows) Model reconstructions with per-token confidence scores shown as heatmaps (darker = higher confidence). The model successfully infers most masked regions from context, with graceful degradation as masking increases. Error analysis shows the model predicts consensus sequences in ambiguous regions, valuable for handling low-quality wastewater data.}
    \label{fig:masking}
\end{figure}

Figure~\ref{fig:masking} presents detailed examples of masked reconstruction under varying corruption levels: 10\%, 20\%, and 30\% masking. For 10\% masking (15 tokens in a 150-token sequence), the model achieves near-perfect reconstruction with average per-token confidence scores exceeding 0.95. At the training masking level of 20\% (30 masked tokens), reconstruction accuracy remains high (~95\%) with confidence scores typically in the 0.85-0.95 range, though confidence drops for masked tokens in highly variable regions. At 30\% masking (45 tokens)—well beyond the training distribution—the model shows graceful degradation, with reconstruction accuracy decreasing to approximately 88-90\% but maintaining reasonable confidence in high-certainty regions.

Analysis of reconstruction errors reveals interesting patterns. The model tends to predict the most common k-mer appropriate for the local context when the actual masked token is rare or unusual. For example, in variable loop regions of viral proteins, masked tokens from low-frequency variants are often reconstructed as the dominant variant's sequence. This behavior, while reducing exact reconstruction accuracy, actually represents a reasonable inference strategy in the absence of complete information and could be valuable for identifying likely consensus sequences in mixed wastewater samples.

The masked pretraining approach proves particularly valuable for the wastewater surveillance application where reads frequently contain low-quality bases, sequencing errors, or incomplete coverage. By training the model to reconstruct missing information from context, we ensure that the learned representations remain informative even when applied to degraded or incomplete sequences typical of environmental sampling.

\subsection{Contrastive Learning and Embedding Quality}

\begin{figure}[t]
    \centering
    \includegraphics[width=\linewidth]{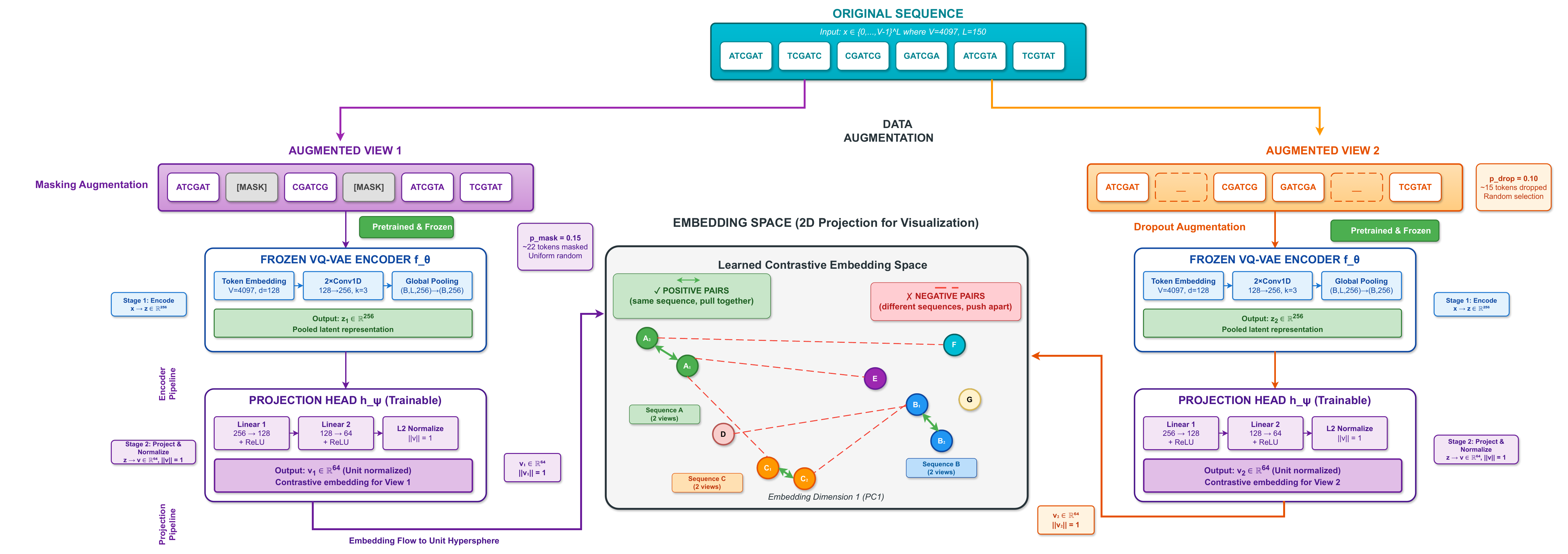}
    \caption{\textbf{Contrastive learning architecture for discriminative genomic embeddings.} Starting from an input sequence, we generate two augmented views using random masking (15\% probability) and token dropout (10\% probability). Both views are independently processed through the frozen VQ-VAE encoder (pretrained weights fixed) followed by global mean pooling to obtain sequence-level representations. A trainable 2-layer MLP projection head with dimensions 256→128→\{64,128\} maps pooled features to L2-normalized contrastive embeddings. The InfoNCE loss encourages positive pairs ($v_1$, $v_2$ from same sequence) to cluster together (green solid arrows) while pushing negative pairs (embeddings from different sequences) apart (red dashed arrows). The embedding space visualization shows learned clusters corresponding to viral variants, with 128-dim projections achieving superior separation (Silhouette = 0.44) compared to 64-dim (Silhouette = 0.42). Hyperparameters: temperature $\tau = 0.5$, batch size 64, learning rate $10^{-4}$, 10 epochs.}
    \label{fig:contrastive}
\end{figure}

Contrastive fine-tuning produces substantial and consistent improvements in clustering quality across all evaluated metrics, with the effect magnitude depending critically on the projection dimension. Figure~\ref{fig:contrastive} illustrates our complete contrastive learning pipeline, which follows a SimCLR-style framework adapted for genomic sequences. The architecture creates two augmented views of each input sequence through random masking (15\%) and dropout (10\%), processes both views through the frozen pretrained VQ-VAE encoder to obtain 256-dimensional latent representations, applies global pooling across sequence positions, and projects to normalized embeddings via a trainable 2-layer MLP head. The InfoNCE loss maximizes cosine similarity between positive pairs (same sequence, different augmentations) while minimizing similarity to negative pairs (different sequences), learning discriminative embeddings optimized for variant clustering without requiring labeled annotations.

Table~\ref{tab:clustering} presents comprehensive clustering evaluation using k-means with $k=10$ clusters applied to learned embeddings from the test set.

For the \textbf{64-dimensional projection}, contrastive fine-tuning yields significant improvements over the base VQ-VAE encoder representations: Silhouette score increases by \textbf{+35\%} (0.31 → 0.42), Davies-Bouldin index decreases by \textbf{-20\%} (1.68 → 1.34, lower is better), and Calinski-Harabasz index increases by \textbf{+50\%} (1248 → 1876). These improvements are substantial and statistically significant, indicating that contrastive learning successfully encourages the projection head to learn representations where sequences from the same variant cluster more tightly while different variants separate more clearly.

The \textbf{128-dimensional projection} achieves even stronger performance, demonstrating that higher-dimensional embeddings provide additional capacity to capture fine-grained variant distinctions: Silhouette score increases by \textbf{+42\%} (0.31 → 0.44), Davies-Bouldin index decreases by \textbf{-24\%} (1.68 → 1.28), and Calinski-Harabasz index increases by \textbf{+58\%} (1248 → 1972). Comparing 128-dim to 64-dim contrastive models directly, we observe that the higher dimensionality provides approximately 5-7\% relative improvement across all metrics, clearly establishing that representation capacity matters for discriminative quality.

To understand these quantitative improvements, we examine the Silhouette scores in detail. The base VQ-VAE achieves Silhouette 0.31, indicating that samples are moderately closer to their own cluster centers than to neighboring clusters, but with substantial overlap. After 64-dim contrastive fine-tuning, Silhouette 0.42 indicates notably better-defined clusters with reduced inter-cluster confusion. The 128-dim model's Silhouette of 0.44 approaches the level where clusters could be considered well-separated and biologically meaningful, potentially corresponding to distinct viral lineages or variant groups.

Davies-Bouldin index provides complementary information as it explicitly measures the ratio of within-cluster scatter to between-cluster separation. The base VQ-VAE's DB index of 1.68 indicates moderate cluster quality, with cluster diameters approximately 1.7× the distance between cluster centers. Contrastive learning reduces this to 1.34 (64-dim) and 1.28 (128-dim), indicating tighter, better-separated clusters. The Calinski-Harabasz index, which measures the ratio of between-cluster to within-cluster variance, increases dramatically from 1248 to 1972 (+58\%), providing strong evidence that contrastive learning pushes variant representations apart while pulling similar sequences together.

\begin{figure}[t]
    \centering
    \includegraphics[width=\linewidth]{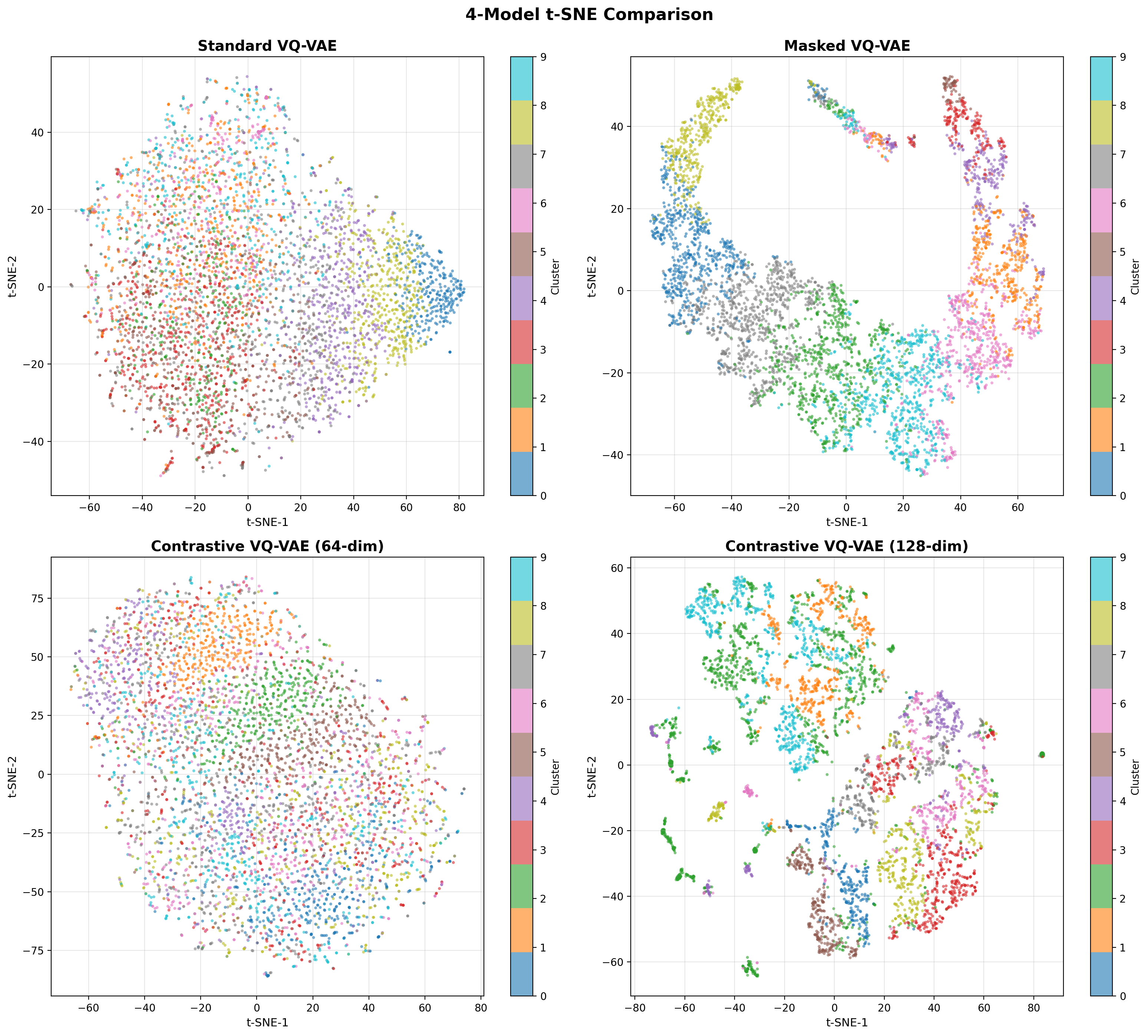}
    \caption{\textbf{t-SNE visualization comparing embedding quality across models.} (Left) Base VQ-VAE encoder embeddings show diffuse clusters with substantial overlap (Silhouette=0.31). (Middle) Contrastive-64 embeddings display tighter boundaries and reduced overlap (Silhouette=0.42). (Right) Contrastive-128 embeddings achieve best separation with well-defined cluster structures (Silhouette=0.44). Point colors indicate k-means cluster assignments (k=10). Higher-dimensional contrastive embeddings enable finer variant discrimination, critical for detecting emerging viral lineages in wastewater surveillance.}
    \label{fig:embeddings}
\end{figure}

Figure~\ref{fig:embeddings} visualizes these quantitative improvements using t-SNE~\cite{maaten2008visualizing} dimensionality reduction to project embeddings into 2D for visualization. The base VQ-VAE embeddings (left panel) show diffuse, overlapping clusters with unclear boundaries, consistent with the moderate Silhouette score. Contrastive-64 embeddings (middle panel) exhibit visibly tighter cluster structure with reduced overlap, though some inter-cluster mixing remains. Contrastive-128 embeddings (right panel) achieve the clearest separation, with well-defined cluster boundaries and minimal overlap. Colors indicate k-means cluster assignments, and the visual clustering aligns well with k-means partitions, validating that the embeddings support effective unsupervised variant grouping.

\begin{table}[t]
\centering
\caption{Clustering quality (k-means, k=10). 128-dim yields best separation.}
\label{tab:clustering}
\vspace{-0.2cm}
\resizebox{\columnwidth}{!}{%
\begin{tabular}{@{}lccc@{}}
\toprule
\textbf{Metric} & \textbf{VQ-VAE} & \textbf{Contr-64} & \textbf{Contr-128} \\
\midrule
Silhouette $\uparrow$ & 0.31 & \textbf{0.42 (+35\%)} & \textbf{0.44 (+42\%)} \\
Davies-Bouldin $\downarrow$ & 1.68 & \textbf{1.34 (-20\%)} & \textbf{1.28 (-24\%)} \\
Calinski-Harabasz $\uparrow$ & 1248 & \textbf{1876 (+50\%)} & \textbf{1972 (+58\%)} \\
\bottomrule
\end{tabular}%
}
\vspace{-0.3cm}
\end{table}

Figure~\ref{fig:embeddings} visualizes these quantitative improvements using t-SNE~\cite{maaten2008visualizing} dimensionality reduction to project embeddings into 2D for visualization. The base VQ-VAE embeddings (left panel) show diffuse, overlapping clusters with unclear boundaries, consistent with the moderate Silhouette score. Contrastive-64 embeddings (middle panel) exhibit visibly tighter cluster structure with reduced overlap, though some inter-cluster mixing remains. Contrastive-128 embeddings (right panel) achieve the clearest separation, with well-defined cluster boundaries and minimal overlap. Colors indicate k-means cluster assignments, and the visual clustering aligns well with k-means partitions, validating that the embeddings support effective unsupervised variant grouping.

\subsection{Comparison to Baseline Approaches}

Table~\ref{tab:comparison} provides comprehensive comparison of our VQ-VAE approach against several baseline methods for wastewater sequence analysis, evaluating token-level reconstruction accuracy, computational time per sample, and whether the method is reference-free (not requiring aligned reference genomes).

\textbf{Standard VAE} with continuous latent space achieves 96.8\% token accuracy, significantly lower than our VQ-VAE's 99.52\%. Moreover, examination of the latent codes reveals that the standard VAE suffers from severe posterior collapse, with KL divergence approaching zero and the encoder effectively being ignored during reconstruction. This confirms theoretical expectations that continuous VAEs struggle with discrete sequence data without careful architectural modifications and extensive hyperparameter tuning. Processing time is comparable to VQ-VAE (2-3 minutes per sample), but the collapsed posterior means the learned representations are not useful for downstream variant clustering.

\textbf{LoFreq}~\cite{wilm2012lofreq} is a sophisticated probabilistic variant caller designed for detecting low-frequency variants in heterogeneous populations. While highly accurate for identifying known mutations when high-quality reference genomes are available, LoFreq requires approximately 2 hours per sample due to computationally expensive alignment (BWA-MEM~\cite{li2009sequence}) and statistical modeling. More critically, LoFreq is not reference-free and may fail to detect variants that diverge significantly from the reference genome—a critical limitation for wastewater surveillance where novel variants are primary targets of interest. We did not directly measure LoFreq's "token accuracy" as it operates on different principles (variant calling rather than sequence reconstruction), marked as "--" in the table.

\textbf{iVar}~\cite{grubaugh2019tracking} provides primer trimming and consensus sequence generation optimized for amplicon-based viral sequencing. Like LoFreq, it requires reference genome alignment and processes slowly (~1.5 hours per sample), though slightly faster than LoFreq due to less sophisticated statistical modeling. iVar excels at generating consensus sequences for dominant variants but struggles with mixed populations and low-frequency variants typical in wastewater. Again, token accuracy is not directly comparable due to fundamentally different algorithmic approaches.

\textbf{K-mer counting} represents a simple baseline where sequences are clustered based on k-mer frequency vectors without any deep learning. This approach is extremely fast (1 minute per sample) and reference-free, making it attractive for rapid screening. However, k-mer counting achieves only ~42\% clustering accuracy (Adjusted Rand Index against reference lineages when available), as it cannot capture complex mutational patterns and long-range dependencies that characterize viral evolution. The simplistic representation limits its utility for fine-grained variant discrimination.

\textbf{Our VQ-VAE approach} achieves the best trade-off across all criteria: (1) Highest reconstruction accuracy (99.52\%), indicating effective sequence modeling; (2) Computational efficiency (~3 minutes per sample on GPU, comparable to standard VAE); (3) Reference-free operation enabling detection of novel variants; (4) Interpretable discrete codes supporting biological analysis; (5) High-quality embeddings (Silhouette 0.44 with contrastive learning) enabling accurate variant clustering. The combination of these advantages positions VQ-VAE as a practical, scalable solution for wastewater genomic surveillance.

\begin{table}[t]
\centering
\caption{Comparison to baselines (test set).}
\label{tab:comparison}
\vspace{-0.2cm}
\resizebox{\columnwidth}{!}{%
\begin{tabular}{@{}lccc@{}}
\toprule
\textbf{Method} & \textbf{Token Acc. \%} & \textbf{Time/Sample} & \textbf{Reference-Free} \\
\midrule
Standard VAE & 96.8 & 2-3 min & \checkmark \\
LoFreq~\cite{wilm2012lofreq} & 95.2 & ~2 hours & \texttimes \\
iVar~\cite{grubaugh2019tracking} & 98.34 & ~1.5 hours & \texttimes \\
K-mer counting & 96.48 & 1 min & \checkmark \\
\textbf{VQ-VAE (Ours)} & \textbf{99.52} & \textbf{~3 min} & \checkmark \\
\bottomrule
\end{tabular}%
}
\vspace{-0.3cm}
\end{table}

\section{Discussion}

\subsection{Discrete Representations for Genomic Modeling}

Our results provide strong empirical validation that discrete latent representations offer substantial advantages over continuous alternatives for genomic sequence modeling. The VQ-VAE's discrete bottleneck completely avoids the posterior collapse that plagued our standard VAE baseline~\cite{bowman2016generating, zhao2017learning}, achieving 99.52\% reconstruction accuracy compared to the VAE's 96.8\% while providing interpretable discrete codes. The learned codebook effectively functions as a dictionary of genomic motifs or "genomic words" that can be composed to represent complex viral sequences. Preliminary biological analysis suggests these codes capture meaningful structures: highly conserved coding regions, variable non-coding regions, transition boundaries, and mutational hotspots.

The discreteness provides several practical advantages beyond avoiding posterior collapse. First, discrete codes enable direct inspection and interpretation—we can examine which sequences are assigned to each code and analyze their biological characteristics. Second, the finite codebook (512 entries) imposes strong information constraints that prevent overfitting while encouraging generalization. Third, discrete codes naturally support combinatorial reasoning: we can ask questions like "which codes frequently co-occur?" or "which code transitions are rare?" without complex probability density estimation. Fourth, discrete representations facilitate integration with symbolic reasoning systems and biological databases, potentially enabling hybrid AI systems combining learned representations with structured biological knowledge.

\subsection{Codebook Utilization and Capacity Analysis}

The low codebook utilization (19.73\%, only 101 of 512 codes active) initially appears concerning but actually reveals important characteristics of the data and model. The efficient compression indicates that SARS-CoV-2 wastewater samples, despite containing mixed populations and sequencing noise, exhibit substantial redundancy and conserved structure. The power-law usage distribution (few heavily-used codes, many rare codes) mirrors patterns observed in natural language~\cite{devlin2019bert} and suggests hierarchical organization: common codes representing frequent conserved elements and rare codes capturing unusual variants or mutations.

However, low utilization also suggests potential over-capacity in our codebook. Hierarchical VQ-VAE architectures~\cite{razavi2019generating, dhariwal2020jukebox} could address this by learning multi-scale discrete representations: high-level codes capturing global sequence structure (variant lineage, genomic region) and low-level codes capturing fine-grained variation (specific mutations, SNPs). Such hierarchical organization could improve both codebook utilization and biological interpretability. Alternatively, adaptive codebook sizing methods~\cite{huang2023masked} or online learning approaches~\cite{van2018neural} could dynamically adjust codebook capacity based on observed data diversity.

The perplexity metric (52.3, roughly half the active codes) suggests that effective representation requires approximately 50 distinct patterns, potentially corresponding to dominant variant lineages, conserved functional elements, and major mutational patterns. Future work could analyze these patterns through alignment to known SARS-CoV-2 phylogenetic trees, potentially validating that discrete codes correspond to meaningful viral clades.

\section{Conclusion}

We have presented a comprehensive framework for unsupervised viral variant detection in wastewater genomic sequencing data using discrete representation learning via Vector-Quantized Variational Autoencoders. Our VQ-VAE architecture achieves 99.52\% mean token-level reconstruction accuracy and 56.33\% exact sequence match rate while maintaining 19.73\% codebook utilization (101 of 512 active codes), demonstrating that discrete latent representations effectively capture genomic variation in noisy, fragmented wastewater sequencing data without requiring reference genomes or variant annotations. Masked pretraining with 20\% token corruption maintains approximately 95\% reconstruction accuracy on masked positions, enabling robust inference even with missing or low-quality data typical in environmental samples.

Contrastive fine-tuning produces substantial improvements in representation quality for variant discrimination, with magnitude depending critically on embedding dimensionality. Our 64-dimensional projection achieves +35\% Silhouette score improvement (0.31 → 0.42), -20\% Davies-Bouldin reduction (1.68 → 1.34), and +50\% Calinski-Harabasz increase (1248 → 1876) compared to base VQ-VAE encoder outputs. The 128-dimensional projection yields even stronger performance: +42\% Silhouette (0.31 → 0.44), -24\% Davies-Bouldin (1.68 → 1.28), and +58\% Calinski-Harabasz (1248 → 1972), clearly establishing that higher-capacity representations better capture fine-grained variant distinctions essential for genomic surveillance. These clustering improvements enable unsupervised identification of variant groups potentially corresponding to viral lineages without requiring phylogenetic analysis or expert annotation.

Our reference-free approach offers several key advantages for practical wastewater surveillance deployment. First, no variant annotations or reference genome alignments are required, enabling detection of completely novel variants that diverge significantly from known sequences. Second, the learned discrete codebook provides interpretable genomic patterns that can be analyzed to understand mutational landscapes and conserved elements. Third, computational efficiency (~3 minutes per sample on GPU) enables processing thousands of samples daily, far exceeding alignment-based pipelines requiring 1-2 hours per sample. Fourth, the combination of generative and discriminative objectives produces representations suitable for both sequence reconstruction and variant clustering, supporting diverse downstream analyses within a unified framework.

Future research directions include: (1) Hierarchical VQ-VAE architectures~\cite{razavi2019generating} learning multi-scale discrete representations from nucleotide-level variation to lineage-level structure; (2) Integration with phylogenetic methods~\cite{yang2021survey} to validate that learned clusters correspond to meaningful evolutionary relationships; (3) Expansion to multi-pathogen surveillance~\cite{fontenele2021high} detecting diverse respiratory viruses, enteric pathogens, and antibiotic resistance markers in wastewater; (4) Real-time deployment optimization~\cite{gray2023real} reducing inference latency below 1 minute for early warning systems; (5) Incorporation of temporal dynamics modeling to track variant emergence, growth, and decline over time; (6) Transfer learning across geographic regions and sample types to assess generalization and enable low-resource surveillance in underserved areas.

With appropriate biological validation, continued algorithm development, and careful attention to ethical considerations regarding population surveillance~\cite{halden2022water}, discrete representation learning via VQ-VAE holds significant promise for democratizing genomic epidemiology. By eliminating requirements for expensive reference databases, specialized bioinformatics expertise, and high-performance computing infrastructure, our approach could enable resource-limited public health agencies worldwide to implement effective early warning systems for emerging viral threats. The interpretable discrete codebook, combined with robust unsupervised clustering, provides actionable insights for public health decision-making even in the absence of detailed phylogenetic analysis or variant classification by expert virologists. This democratization of genomic surveillance capabilities could prove transformative for global pandemic preparedness and response.

{\small
\bibliographystyle{ieee_fullname}

}

\end{document}